\documentclass[12pt]{elsarticle}

\biboptions{numbers,sort&compress}
\usepackage{graphicx}
\usepackage{amsmath,amssymb,amsfonts}
\usepackage{subcaption}
\usepackage{hyperref}
\hypersetup{
    colorlinks=true,
    linkcolor=blue,
    filecolor=blue,
    urlcolor=blue,
    citecolor=blue
}

\begin{document}

\title{Efficient and Distributed Large-Scale \\
3D Map Registration using Tomographic Features}

\author[1]{Halil Utku Unlu}
\author[2,3]{Anthony Tzes}
\author[1,3]{Prashanth Krishnamurthy}
\author[1,3]{Farshad Khorrami}

\affiliation[1]{organization={Electrical \& Computer Engineering Department},
            addressline={6 MetroTech Center}, 
            city={Brooklyn},
            postcode={11201}, 
            state={New York},
            country={USA}}
\affiliation[2]{organization={Electrical Engineering, New York University Abu Dhabi (NYUAD)},            addressline={Saadiyat Island}, 
            postcode={129188}, 
            state={Abu Dhabi},
            country={UAE}}
\affiliation[3]{organization={Center for Artificial Intelligence and Robotics, NYUAD},
            addressline={Saadiyat Island}, 
            postcode={129188},
            state={Abu Dhabi},
            country={UAE}}

\begin{abstract}
A robust, resource-efficient, distributed, and minimally parameterized 3D map matching and merging algorithm is proposed. The suggested algorithm utilizes tomographic features from 2D projections of horizontal cross-sections of gravity-aligned local maps, and matches these projection slices at all possible height differences, enabling the estimation of four degrees of freedom in an efficient and parallelizable manner. The advocated algorithm improves state-of-the-art feature extraction and registration pipelines by an order of magnitude in memory use and execution time. Experimental studies are offered to investigate the efficiency of this 3D map merging scheme.
\end{abstract}
\begin{keyword}
Map merging, 3D perception, Multi-robot exploration
\end{keyword}

\maketitle

\section{Introduction}\label{sec:introduction}

In many applications, robot collaboration not only is required, but also greatly enhances the effectiveness of a robotic system. Multiple robots can provide a larger coverage \cite{tzes2018visual}, different views of the working environment, faster exploration, and resiliency against individual agent failures. However, communication fidelity, limited computational capabilities, and the operated environments' dynamic and uncertain nature hamper the utility of multi-agent systems in real-world applications \cite{lajoie2022towards}.

A baseline requirement for a collaborative mission for robotic systems is a shared understanding of the environment and their location within it. External infrastructures with a centralized computation structure can provide the global state to each agent, but they restrict the applicability of the systems in unknown, unstructured environments without a central unit. 

The common solution for autonomous localization, navigation, and mapping in unknown, uncertain environments is the set of methods called simultaneous localization and mapping (SLAM), in which the robotic platforms rely on the data they collect through onboard sensors to estimate their position and generate a map. Single robot SLAM for both 2D and 3D motion is a mature field with many advanced platforms and frameworks~\cite{kazerouni2022survey, huang2023indoor}.

An extension of the SLAM methods on multiple agents is commonly referred to as Collaborative SLAM (C-SLAM), and it aims to provide a consistent and more accurate system state estimate (i.e., location of all agents, global map) by utilizing data from multiple participating agents. The field has seen significant development in recent years, though large-scale deployment needs further refinement~\cite{lajoie2022towards}.
Part of the reason is that the SLAM task requires the use of dense, rich data from vision sensors to generate a useful and accurate environment representation. 3D data streams (RGB-D sensors and LiDAR) generate large amount of data requiring a high communication bandwidth that can overload the network and demand more computational budget from the resource-limited mobile agents.

An alternative way for utilizing individual map information gathered by every agent in the field is to perform map matching and merging, where the agents share their map with other agents upon connectivity and attempt to align and combine other maps, hence providing a more complete view of their environment. Multiple approaches have been proposed for 2D map matching and merging scenarios~\cite{carpin2005map, saeedi2014map, ding2019deepmapping, andersone2019heterogeneous}, but the solutions do not easily scale to 3D map representations. Although the advances in low-power devices with GPUs address partially the computation issue, the complexity needed from the robotic agents adds additional burdens on already resource-strapped platforms. Furthermore, many of the existing algorithms require adjusting the parameters of the network, either through training on a different data or adjusting a large set of key parameters. 

This paper attempts to address the gap in map matching for large-scale 3D maps by proposing a feature extraction framework that effectively enables the use of 2D image features in a pair of gravity-aligned 3D maps using tomography~\cite{unlu2023distributed} and providing a thorough evaluation and comparisons against alternatives. In the proposed method, the algorithm a) extracts a binary occupancy representation for horizontal cross-sections of the maps, b) computes 2D features over these slices, and c) restricts matching space for the features across maps to its corresponding section only. Cross-correlating the maps based on their agreement in pose estimations yields a 4 degree of freedom (DoF) transformation estimate. The proposed approach will be demonstrated to be significantly efficient (in both time and memory consumption) and accurate compared to state-of-the-art point cloud registration methods.
An early version of this paper appeared in~\cite{unlu2023distributed}, which does not contain the simulation studies, any experimental results, and the recent algorithms over the past two years.

The contributions of this article include:
\begin{itemize}
    \item a simple and efficient approach to extracting 3D features via tomographic extraction of horizontal sections with minimal parameterization,
    \item study of viable uses of the aforementioned features in addressing large-scale 3D-map matching and merging scenarios across environments of varying scales,
    \item extensive comparative studies on simulated and real data to assess effectiveness against alternative methods, 
    \item experimental studies using a quadruped to emulate merging scenario in a large indoor area, and
    \item the source code to replicate the study for both proposed and alternative methods\footnote{\href{https://github.com/RISC-NYUAD/tomographic-map-matching}{\texttt{https://github.com/RISC-NYUAD/tomographic-map-matching}}}.
\end{itemize}

The remainder of the paper is structured as follows: Relevant work is introduced in Section~\ref{sec:related-work}, and the formulation of the studied problem is provided in Section~\ref{sec:problem-formulation}. The proposed method is described in Section~\ref{sec:proposed-method}, detailing the process of tomographic feature extraction and registration with the computed features.
The structure of the comparative analysis and the baseline methods used for evaluation are detailed in Section~\ref{sec:comparative_studies}. Results on a simulated dataset are illustrated in Section~\ref{sec:simulation-data}. Experimental studies on real data, using KITTI odometry benchmark and a newly generated data from a large-scale indoor environment, are provided in Section~\ref{sec:experimental_studies}. Limitations of the algorithm and the overall study are provided in Section~\ref{sec:limitations}, followed by the concluding remarks.

\section{Related Work}\label{sec:related-work}
\subsection{Map Matching and Merging}

Map \emph{matching} is related to estimating the relative transformation between a pair of maps, and map \emph{merging} is the process of combining the representations of a set of maps to a common structure, given the relative transformation. The literature for matching and merging probabilistic 2D occupancy grid maps is vast and several reviews provide a thorough coverage of the field \cite{andersone2019heterogeneous, yu2020review}. Overall, two common threads can be observed: optimization-based and feature-based map matching.

Optimization-based algorithms operate on the entire probability grid without an intermediate representation. Some of the earlier work focused on maximizing a similarity metric \cite{carpin2005map} and a connectivity metric with better results \cite{erinc2014anytime}. More recent work uses genetic algorithms to address the matching and merging on low-resource platforms \cite{sun2023method}. Feature-based map matching algorithms extract an intermediate representation from the raw occupancy grid to perform associations and estimations. Salient and descriptive feature extraction forms the basis of matching \cite{aragues2011distributed, blanco2013robust, lee2020tomographic, chen2021map, wang2023multi}.

Tomographic features and a 2D map matching based on the Radon transform of the occupancy grid is employed in~\cite{lee2020tomographic}. Tomographically salient features are used in a Gaussian mixture model to estimate the 3 DoF transformation. Another feature-based algorithm extracts 2D image features from image representations of the occupancy grid maps and performs an initial guess via RANSAC, which is then refined with ICP for tighter alignment~\cite{blanco2013robust}. Extensive comparison of various image features and ideal parameter selection process is provided in the paper.

The algorithms outlined earlier cannot be applied in 3D domain directly. The main issues in 3D map matching stem primarily from the scale and density of data with the added dimension \cite{xie2020map}. 3D LiDAR sensors provide accurate position information in a long range, while stereo cameras with onboard computational units can measure dense depth in a camera field of view. Maps generated from such sensors are inevitably large. The ideas for matching and merging are similar to those in the 2D case: optimize a distance cost over 3D data \cite{bonanni20173, mohanarajah2015cloud}, or extract local or global features from the 3D data and perform a robust registration using the corresponding features \cite{yue2018hierarchical, basso2023merging, yin2023automerge, stathoulopoulos2023frame}.

However, the aforementioned methods cannot run in real time using onboard computer of the individual agents. Near real-time performance was achieved in \cite{yue2018hierarchical} while providing demonstrations on large scale but simplistic indoor environments. Similarly \cite{stathoulopoulos2023frame} relies on learning-based place recognition descriptors from pose-centered spherical projections and GICP optimization, but require preprocessing on each scan to be performed. While the SLAM and map matching and merging operations both require the preprocessing step, the systems end up being coupled.

\subsection{Collaborative SLAM}

The primary focus for collaborative SLAM (C-SLAM) algorithms \cite{lajoie-2022-towar-collab}, is the pose optimization using both intra-robot and inter-robot loop closures within the maps. As such, individual agents do not share their maps fully with others in the network. With the exception of some centralized architectures \cite{schmuck2019ccm, Chang2022LAMP2A}, it is not possible to obtain the global map with many of the prominent C-SLAM architectures \cite{lajoie2020door, tian2022kimera, Lajoie2023SwarmSLAMS}. It is indeed possible to perform the map merging using the C-SLAM algorithms, but a framework for sharing map data between agents is not discussed, and the bandwidth requirements for dense 3D map data remains a significant challenge.

\subsection{3D Point Features}

A common issue that needs to be addressed for both 2D and 3D feature description problem is the necessity for the descriptors to be pose invariant. The same point of interest, viewed from different locations, should result in ideally the same descriptor. An additional challenge for arbitrary 3D point clouds is that the data structure is not necessarily ordered, and any permutation of the order of points should still provide accurate descriptors. Much of the work is focusing on obtaining such pose-invariant descriptors, and can be grouped into two major categories: hand-crafted and learning-based.

Hand-crafted feature descriptors establish a local frame of reference, informed by the point neighborhood, to generate discerning statistics about the point. A prominent algorithm is Fast Point Feature Histograms (FPFH)~\cite{Rusu2009FastPF}, comprised of the histograms of angular variations for a point of interest in its k-nearest neighborhood. A comprehensive review of hand-crafted features in~\cite{Guo2015ACP} found FPFH to be effective in low-noise scan matching scenarios. An adaptation of FAST \cite{trajkovic1998fast} corner detection on orthogonal 2D projections of the neighborhood of a point is proposed as an efficient alternative for low-resource platforms \cite{basso2023merging}.

Recent algorithms attempt to learn the discerning and pose-invariant descriptors directly from available object, indoor, and outdoor datasets. Earlier realizations relied on adapting Convolutional Neural Networks (CNNs) to 3D voxel-based representations \cite{wu20153d, Zeng20163DMatchLL}. Structured use of multi-layer perceptrons were shown to be effective in dealing with the unordered nature of point cloud data \cite{qi2017pointnet, Deng2018PPFNetGC}, and various CNN-architectures that focus directly on point data emerged~\cite{Hua2017PointwiseCN, Atzmon2018PointCN, Gojcic2018ThePM, Bai2020D3FeatJL, floris2024composite}. 

A review of learning-based point cloud feature extraction for various applications can be found in \cite{zhang2020deep}. Two of the convolutional architectures are important to our study: Kernel Point Convolution (KPConv)~\cite{Thomas2019KPConvFA}, which uses radius-neighborhoods to define deformable kernels to be used in 3D point convolution, and Fully Convolutional Geometric Features (FCGF)~\cite{Choy2019FullyCG}, which implements a fully-convolutional architecture with a metric learning loss to improve both accuracy and efficiency.

\subsection{Point Cloud Registration}

Given noiseless data and perfect correspondences, the registration problem has a closed form solution by solving the Procrustes problem. However, the conditions cannot be met in real world. Sensors are noisy, and correspondences can be wrong: 95\% outlier rates have been reported \cite{bustos2017guaranteed}, necessitating robust solutions. Reviews for the algorithms addressing registration problem for same-source \cite{huang2021comprehensive} and cross-source \cite{huang2023cross} are available.

Sample consensus-based algorithms (SAC or RANSAC-family) \cite{fischler1981random} find the most consistent registration hypothesis by evaluating the solution generated from a minimal set of correspondences, sampled from the initial correspondence set, and finding the solution that is accurate for the most number of correspondences, termed the \textit{inlier set}. More modern iterations of RANSAC family of algorithms provide optimality guarantees in quadratic-time \cite{li2023qgore} and differentiable implementation for trainability and use in learning-based frameworks \cite{wei2023generalized}. Another significant robust registration method relies on semi-definite relaxation in its optimization, allowing it to handle extreme outlier ratios with a certificate of optimality \cite{yang2020teaser}. 

Learning-based registration frameworks cast the problem into differentiable sets of modules, treating initial set of correspondences as putative and assigning weights on each correspondence. Deep Global Registration (DGR)~\cite{choy2020deep} minimizes a robust energy metric with a convolutional network for confidence assignment, while PointDSC~\cite{bai-2021-point} incorporates spatial consistency explicitly to the network architecture.

Attention-based transformer networks aim to utilize the benefits of transformer architectures for point cloud registration problems. Some examples include PREDATOR~\cite{huang-2021-predat}, a model that learns to focus primarily on the overlapping regions, GeoTransformer \cite{qin2023geotransformer}, an archirecture that encodes the geometric structure of the point neighborhood to obtain a pose-invariant superpoint representation and performs keypoint-free coarse-to-fine registration, and Rotation-Invariant Transformer for Point Cloud Matching (RoITr) \cite{yu2023rotation}, a system trained to be able to deal with arbitrary rotational transformations.

A significant bottleneck for the attention mechanisms is the computational complexity. Most of the algorithms appear to be implemented and tested only on server-grade GPUs, without much concern for mobile deployment. BUFFER \cite{ao2023buffer} stands out as a model that attempts a trade-off between accuracy, efficiency, and generalizability by improving the feature quality and estimation pipelines with small-scale networks.

\section{Problem Formulation}\label{sec:problem-formulation}

Vectors and matrices are denoted with lowercase and uppercase boldfaced characters (e.g. $\mathbf{p}$, $\mathbf{T}$), respectively. Sets are named with calligraphic fonts (e.g. $\mathcal{S}$, $\mathcal{C}$) or otherwise implicitly defined inside curly brackets. The cardinality of a set is denoted as $|\cdot|$. $\mathbb{R}$ and $\mathbb{Z}^+$ denote the set of real numbers and positive integers, respectively. Object indexing and memberships are denoted with a combination of pre- or post-fixed  subscripts and superscripts.

\subsection{Map Matching Problem}

This paper addresses the problem of pairwise merging of gravity-aligned 3D maps that can be canonically represented as an unordered point cloud, where each point stores a coordinate as ${^i\mathbf{p}_j} \in \mathbb{R}^3,~i \in \{c, d\}$, by two agents $c$ and $d$. We assume no prior knowledge of the absolute pose information for any agent, and the agents are not required to observe each other via a rendezvous. The communication network is assumed to have sufficient bandwidth in the order of 10~Mbps in order to transfer incremental maps having a size close to 70~MB. The map matching and merging tasks are invoked upon sufficient connectivity.

The map of different agents will be denoted as
\begin{equation}
    \label{eq:map-definition}
    {^i\mathcal{M}} = \{\dots,~{^i\mathbf{p}_j},~\dots  \},
    \quad |{^i\mathcal{M}}| = N_i,
    \quad i \in \{c,~d\}.
\end{equation}
To ensure the applicability and scalability of the tested algorithms, it is assumed that the point cloud representation is processed with a voxel grid filter of leaf size $g$. The filtering process ensures more uniform point density and eliminates the dependency on particular sensor sets during mapping (i.e., LiDAR providing high point density perpendicular to its rotation axis, RGBD/stereo having high density in limited field of view).

The number of points in agents' maps, $N_i$, depends primarily on the map resolution and the size of the environment that is being mapped. Sensible combinations that preserve a moderate amount of detail can increase the map size quickly, reaching the order of millions.

Let $w$ denote a frame of reference fixed in the world. The points of the map ${^i\mathcal{M}}$ can be transformed into $w$ via a rigid 3D homogeneous transformation $\mathbf{T}_i^w \in SE(3)$:
\begin{equation}
    \label{eq:lie-group-defn}
    SE(3) = 
    \left\{ 
        \begin{bmatrix}
            \mathbf{R}_{3 \times 3} & \mathbf{t}_{3 \times 1} \\
            \mathbf{0}_{1 \times 3} & 1
        \end{bmatrix}, \quad
        \mathbf{R}^\top\mathbf{R} = \mathbf{RR}^\top = \mathbf{I}, \quad
        \det({\mathbf{R}}) = 1 
    \right\}
\end{equation}

The global world frame $w$ can be arbitrarily defined by anchoring one agent's frame as the map origin. Assigning the local frame of agent $c$ as the fixed frame $w$, the goal is to estimate the transformation $\mathbf{T}_d^c \in SE(3)$.

We assume that the agents' maps are gravity-aligned. More explicitly, we assume that the agents are capable of observing the gravity vector. The gravity vector is common among all of the agents operating in the environment and is parallel to the $z$-axis of the map they are building. We do not restrict the $z$-axes to be identical (i.e., origins in the $z$-axes are coincident), as the agents may have different initializations for the map height.

With this assumption, the problem is restricted to 4 degrees of freedom (DoF) and the homogeneous transformation can be parameterized as
\begin{equation}
    \label{eq:paremeterized-transform}
    \mathbf{T}_d^c = \mathbf{T}(x_d^c, y_d^c, z_d^c, \theta_d^c) =
    \begin{bmatrix}
        \cos \theta_d^c & -\sin \theta_d^c & 0 & x_d^c \\
        \sin \theta_d^c & \cos \theta_d^c & 0 & y_d^c \\
        0 & 0 & 1 & z_d^c \\
        0 & 0 & 0 & 1
    \end{bmatrix}.
\end{equation}
The given setup does not restrict the motion of the agents. The agent is free to execute any motion, as long as the gravity vector can be measured.

A map matching (or more generally, a point cloud registration) algorithm relies on the set of correspondences $\mathcal{C}$ established between pairs of points:
\begin{equation}
    \label{eq:point-pair}
    \mathcal{C} \triangleq 
    \left\{
        ({^c}\mathbf{p}_m, {^d}\mathbf{p}_n):~m, n \in \mathbb{Z}^+,~
        m \leq N_c,~
        n \leq N_d
    \right\}.
\end{equation}

Assuming that all correspondences are ``true", which are the pairs that overlap when the true transformation is applied to the targeted point cloud, the registration task can be cast as an optimization problem as
\begin{equation}
    \label{eq:registration-optim}
    \mathbf{T}_d^c = \underset{\mathbf{T} \in SE(3)}{\arg \min}~\sum_{({^c}\mathbf{p}_m, {^d}\mathbf{p}_n) \in \mathcal{C}} \rho({^c}\mathbf{p}_m, \mathbf{T}~{^d}\mathbf{p}_n)
\end{equation}
where $\rho(\mathbf{a}, \mathbf{b})$ is a non-negative distance metric to determine the error for a given correspondence under the transformation hypothesis $\mathbf{T}$. As an example, in the case of point-to-point iterative closest point (ICP), the cost metric assumes the form of Euclidean distance.

Estimating ``true" correspondences is a difficult task, as it depends on the discrimination capacity of the descriptor and some mismatches cannot be avoided due to perceptual aliasing. It is reasonable to expect outlier correspondence rates of 90\%~\cite{Parra2017GuaranteedOR} for 3D datasets.

An inlier set $\mathcal{C}_{\text{in}} \subseteq \mathcal{C}$ is the set of correspondences that agree with a given transformation, and it needs to be selected from correspondences that agree with the unknown, correct transformation, leading to a modification of \eqref{eq:registration-optim}:
\begin{equation}
    \label{eq:registration-optim-inlier}
    \mathbf{T}_d^c = \underset{\mathbf{T} \in SE(3)}{\arg \min}~\sum_{({^c}\mathbf{p}_m, {^d}\mathbf{p}_n) \in \mathcal{C}_{\text{in}}} \rho({^c}\mathbf{p}_m, \mathbf{T}~{^d}\mathbf{p}_n)~.
\end{equation}

\section{Proposed Method} 
\label{sec:proposed-method}

\subsection{Tomographic Features}
\label{ssec:feature-extraction}

A tomographic section of the map, termed \textbf{slice} throughout the paper, is defined as the 2D binary occupancy representation of a horizontal cross-section of a map $\mathcal{M}_i$ at a particular height $h$. The points in the map to generate a slice at height $h$ can be defined as:
\begin{equation}
    \label{eq:slice-defn}
    {^i\widehat{\mathcal{S}}_h} \triangleq 
    \left\{
    {^i}\mathbf{p}_j:\ h - t < \left.{^i\mathbf{p}_j}\right|_z \leq h + t,
    \right\}
\end{equation}
where $\left.{\mathbf{p}}\right|_z$ is the $z$-coordinate of point $\mathbf{p}$ and $t$ is a parameter that determines the thickness of the cross-section.

A natural choice for the thickness parameter to utilize all available information is $t = g/2$, half the leaf size. In this manner, every point in the map will belong to a single slice for a finite set of $h$ that are $g$ apart. Let $^iz_{\max}$ ($^iz_{\min}$) be the maximum (minimum) $z$-coordinate of the map $^i \mathcal{M}$. Then, the index set of heights $\mathcal{H}_i$ for $^i \mathcal{M}$ is defined as:
\begin{equation}
    \label{eq:slice-indexing}
    \mathcal{H}_i \triangleq 
    \left\{ 
    {^i}z_{\min} + k \cdot g : \ 
    k \in \mathbb{Z}^+, \  
    k \leq \frac{{^i}z_{\max} - {^i}z_{\min}}{g}
    \right\}.
\end{equation}

Points in every slice ${^i\widehat{\mathcal{S}}_h},~h \in \mathcal{H}_i$ are projected onto the $xy$-plane and discretized on a 2D grid to obtain a 2D binary occupancy image, ${^i\mathbf{S}_h}$. The extrema $xy$ coordinates of points in ${^i\widehat{\mathcal{S}}_h}$ and the grid size $g$ dictate the width and height of the resultant binary image. Intuitively, each pixel represents the occupancy of a real area of size $g \times g~m^2$.

\begin{figure}[htbp]
    \centering
    \includegraphics[keepaspectratio, width=\linewidth]{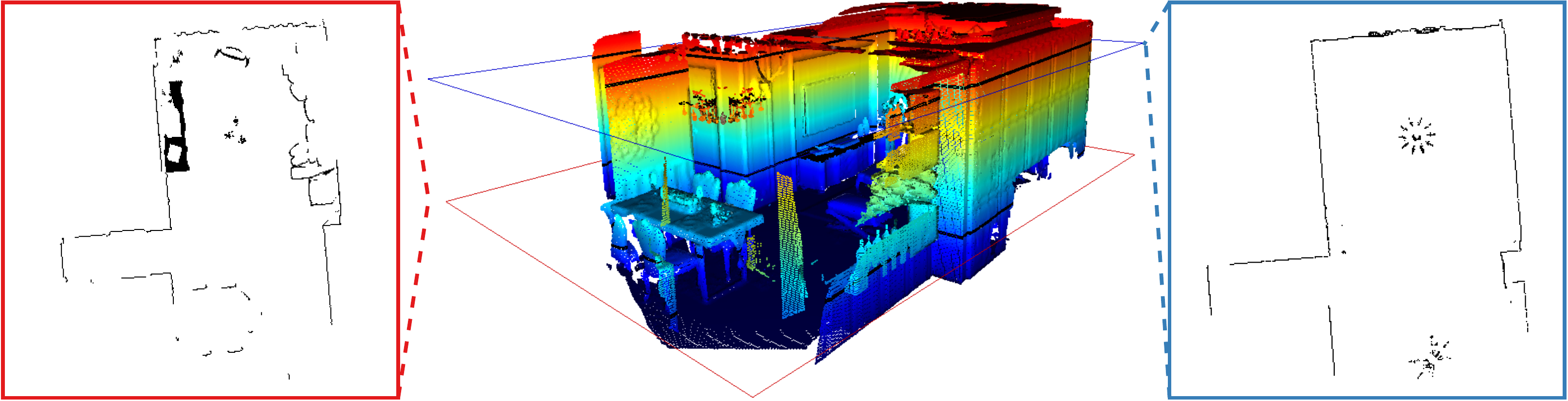}
    \caption{Samples of 2D binary slice images extracted at differnet heights for a simulated indoor environment. Binary images with colorized borders denote the corresponding height at the 3D rendering of the environment.}
    \label{fig:example-slices}
\end{figure}

Examples of the 2D binary images obtained through the slicing are provided in Figure~\ref{fig:example-slices} for an indoor environment.

ORB features and descriptors~\cite{rublee2011orb} are extracted from every 2D binary slice, and the highest scoring $K$ features are kept. The set of 128-bit descriptors, paired with the 2D coordinates, is denoted as ${^i}\mathcal{F}_h$ and forms the basis of the map matching task:
\begin{equation}
    \label{eq:feature-set-defn}
    {^i}\mathcal{F}_h \triangleq \{ {^i}\mathbf{f}_{h, k}\},~ |{^i}\mathcal{F}_h| \leq K.
\end{equation}
The choice of ORB is primarily informed by the lack of real photo intensity changes  within the binary images ${^i\mathbf{S}_h}$ and the efficiency of the underlying FAST~\cite{trajkovic1998fast} corner detection with oriented BRIEF~\cite{calonder2010brief} descriptors. Alternative feature extraction and description pipelines can be used \cite{ma2021image}, though the reliability and efficiency of ORB features are validated in the literature for various cases \cite{tareen2018comparative}.

The slicing and feature extraction results in a different representation of the map as an indexed set of 2D binary slices and associated features:
\begin{equation}
    \label{eq:map-slice-representation}
    \mathcal{S}_i \triangleq 
    \left\{
        ({^i\mathbf{S}_h},~{^i}\mathcal{F}_h),~h \in \mathcal{H}_i
    \right\}.
\end{equation}

\subsection{Map Matching with Tomographic Features}\label{sec:map-matching}

\begin{figure}[htbp]
    \centering
    \includegraphics[keepaspectratio, width=\linewidth]{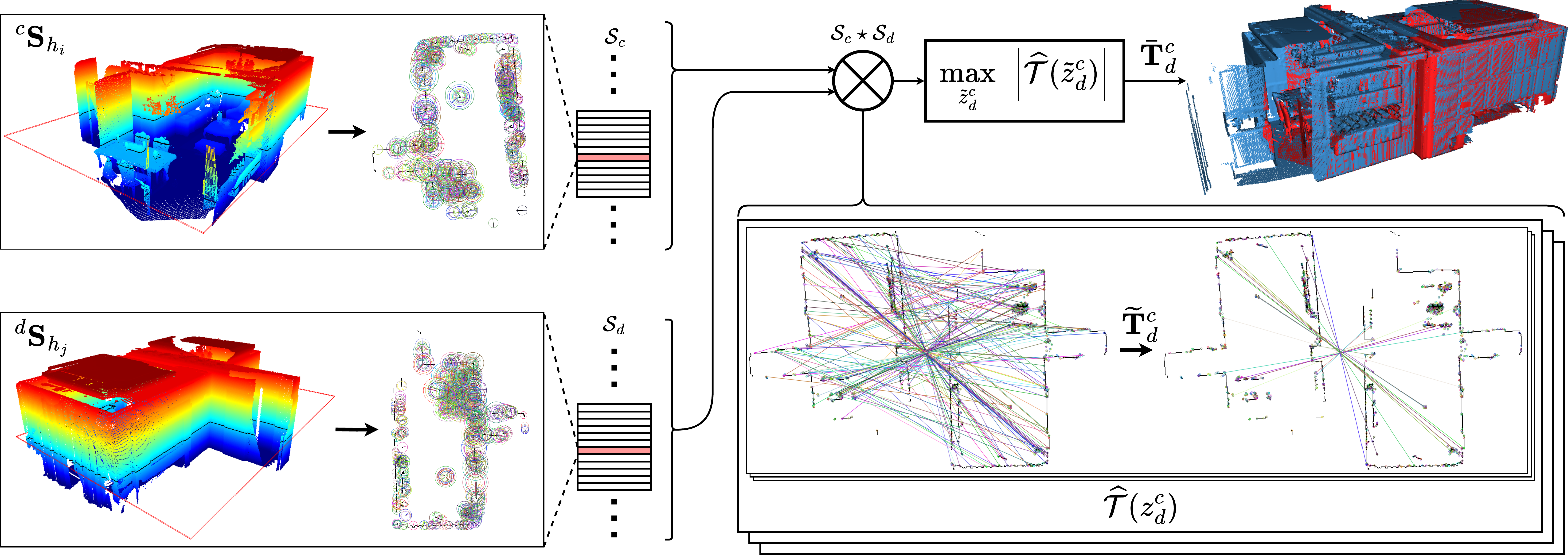}
    \caption{A visual summary of the proposed distributed 3D map matching framework. Each agent $\{c, d\}$ is responsible for extracting slices ${^{\{c, d\}}}\mathbf{s}_{h_i}$ at a predefined grid size. The pose is estimated via two-stage optimization algorithm, where one of the agents cross-correlates the slices by estimating a 3DoF rigid transformation, $\widetilde{\mathbf{T}}_d^c$, between slices, and the consensus of different height hypotheses, $z_d^c$, is used to determine the 4th DoF.}
    \label{fig:visual-summary}
\end{figure}

Given the set of binary slices and their associated features $\mathcal{S}_c$ and $\mathcal{S}_d$, the task is to estimate the 4 unknowns in the transformation $\mathbf{T}_d^c = \mathbf{T}(x_d^c, y_d^c, z_d^c, \theta_d^c)$ (from \eqref{eq:paremeterized-transform}). The proposed method solves a hierarchical optimization problem in two stages: a) the rigid 2D transformation estimation, and b) the height difference estimation, as shown in Figure~\ref{fig:visual-summary}.

\subsubsection{Rigid 2D Transformation Estimation}

The first stage estimates the 3 DoF for different values of $z_d^c$. For a known height difference, each slice in a map will have either one corresponding slice that will be at the same height, or none. For each slice pair $({^c\mathbf{S}_m},~{^d\mathbf{S}_n})_k$ indexed by $k$, matching computed features yields a set of putative correspondences $\{ \cdots, ({^c\mathbf{p}_j},~{^d\mathbf{p}_j}), \cdots \}_k$, and the pose estimation problem reduces to finding a 2D rigid transformation, which has the form
\begin{equation}
    \label{eq:estimate-2d}
    \begin{bmatrix}
        \cos~\theta_d^c & -\sin~\theta_d^c & x_d^c \\
        \sin~\theta_d^c & ~\cos~\theta_d^c & y_d^c
    \end{bmatrix}
    \begin{bmatrix}
        {^dx_j} \\ {^dy_j} \\ 1
    \end{bmatrix}
    =
    \begin{bmatrix}
        {^cx_j} \\ {^cy_j}
    \end{bmatrix}
\end{equation}
for corresponding points ${^i\mathbf{p}_j} = \begin{bmatrix}{^ix_j} & {^iy_j}\end{bmatrix}^\top,~i \in \{c, d\}$. The problem is then cast to a linear system as
\begin{equation}
    \label{eq:linear-problem}
        \begin{bmatrix}
            \vdots & \vdots & \vdots & \vdots \\
            {^dx_j} & -{^dy_j} & 1 & 0 \\
            {^dy_j} & ~{^dx_j} & 0 & 1 \\
            \vdots & \vdots & \vdots & \vdots
        \end{bmatrix}
        \begin{bmatrix}
            \alpha \\ \beta \\ x_d^c \\ y_d^c
        \end{bmatrix}
        =
        \begin{bmatrix}
            \vdots \\ {^cx_j} \\ {^cy_j} \\ \vdots
        \end{bmatrix}
\end{equation}
where $\alpha = s\cos(\theta_d^c)$ and $\beta = s\sin(\theta_d^c)$ with $s$ as the scale parameter. Since we assume known scale, $s=1$. The angle $\theta_d^c$ is recovered using $\text{atan2}(\beta, \alpha)$.

For simplicity, RANSAC is used to obtain a robust solution, followed by Levenberg-Marquardt refinement steps over the inlier set. The linear system given in Equation~\eqref{eq:linear-problem} is solved for every slice pair $k$ at a given height to obtain the set of transformations at a known height:
\begin{equation}
    \label{eq:rigid-estimation-set}
    \widetilde{\mathcal{T}}(z_d^c) \triangleq \left\{
        {_k\widetilde{\mathbf{T}}(\widetilde{x}_d^c, \widetilde{y}_d^c, z_d^c, \widetilde{\theta}_d^c)} \triangleq {_k\widetilde{\mathbf{T}}_d^c}
    \right\}.
\end{equation}

It is not expected that every slice pair $k$ provides an accurate estimate, as the agents may not have complete occupancy information in the neighborhood of a particular $xy$-coordinate at every height. In such cases, the pose estimate ${_k\widetilde{\mathbf{T}}_d^c}$ does not represent the true transform, and simple averaging across all slices will introduce large errors.

To that end, a \emph{consensus} between different 2D rigid estimations is established to obtain the most likely pose estimate. We perform this by finding the largest set of inlier hypotheses $\widehat{\mathcal{T}}(z_d^c) \subseteq \widetilde{\mathcal{T}}(z_d^c)$ within a distance to an anchor hypothesis, ${_k}\widetilde{\mathbf{T}}_d^c$:
\begin{equation}
    \label{eq:per-slice-transform}
    \widehat{\mathcal{T}}(z_d^c) = \arg \max~
    \left\vert
    \left\{ 
    {_k}\widetilde{\mathbf{T}}_d^c:
    \mathbf{d}({_k}\widetilde{\mathbf{T}}_d^c,~{_l}\widetilde{\mathbf{T}}_d^c) \prec \mathbf{t},~
    {_{\{k, l\}}}\widetilde{\mathbf{T}}_d^c \in \widetilde{\mathcal{T}}(z_d^c)
    \right\}
    \right\vert,
\end{equation}
where the vector-valued function $\mathbf{d}$ encodes the Euclidean distance between the estimates of $x_d^c, y_d^c$ and the angular distance between the angles $\theta_d^c$, and $\mathbf{t}$ is a threshold to specify maximum allowed deviations. The parameter-wise average of the hypotheses is used in the inlier set $\widehat{\mathcal{T}}(z_d^c)$ to compute the resultant pose hypothesis, $\bar{\mathbf{T}}_d^c$.

\subsubsection{Height Difference Estimation}

The previous step assumed that the height difference $z_d^c$ between maps is known. Unless there are means to establish the initial height difference, it is a strong assumption that would restrict the applicability of the algorithm to different scenarios.

The $z_d^c$ is similarly estimated based on the consensus. Due to the discretization of the 3D maps via slicing, there are finite number of relative height steps that can be established between a pair maps, all of which are $g$ units apart. Testing all possible height steps can be thought of as ``cross-correlation" of the maps: each step results in a pose hypothesis $\bar{\mathbf{T}}_d^c$ computed from an inlier set $\widehat{\mathcal{T}}(z_d^c)$, and the size of the inlier set is the value for the correlation operator. As such, the height difference $z_d^c$ with the largest cardinality for $\widehat{\mathcal{T}}(z_d^c)$ is selected as the height estimate:
\begin{equation}
    \label{eq:z-estimation}
    z_d^c = \underset{z}{\arg \max}~
    \left\vert \widehat{\mathcal{T}}(z) \right\vert.
\end{equation}

The correct height difference would allow the largest number of slice pairs that themselves agree on the remaining degrees of freedom. Empirical tests indicate that the number of inlier correspondences are maximized for all slices when they are matched with the correct slice from the other map for indoor/outdoor environments with distinctive geometric features. We refer to the proposed method as \textbf{Consensus} throughout the paper.

\subsubsection{Computational Complexity}

The proposed method utilizes various simple algorithms that have a low computational burden individually, but repeated for a large number of hypotheses. Especially, the cross-correlation for maps that span larger heights are computationally expensive. However, each of these individual steps (feature calculation for every slice and 3 DoF estimation for a pair of slices) are independent of each other, enabling parallelization opportunities with hardware acceleration. Moreover, it is not necessary for intermittent map matching and merging on different indoor and outdoor scenarios.

\section{Comparative Studies}
\label{sec:comparative_studies} 

We provide a comprehensive evaluation of the accuracy and efficiency of the proposed algorithm in both simulated and real-life datasets against various registration pipelines from the literature.

\subsection{Algorithm Baselines and Metrics}

Comparisons to eight different baselines are provided. The selection includes standard algorithms commonly used as a baseline in literature, as well as the state-of-the-art registration algorithms. The baselines that are not learning-based study different conditions in feature extraction and registration, and a literature implementation may not necessarily exist for them.

For the learning-based algorithms, the selection was informed primarily on their adaptability to the map matching problem. In all of the cases, the weights provided by the original authors are used. Since some methods do not fit the memory of mobile-grade GPUs, the evaluations are done on a high-performance computer with server-grade GPU (NVIDIA A100 80~GB).

\subsubsection{Tomographic TEASER++}

Tomographic TEASER++ \cite{yang2020teaser} replaces the consensus-based selection of the height difference with a 3D registration problem at each height step, and selecting the solution that provides the largest number of inlier features.

Conversion from 2D ORB features to 3D features are performed by assigning the height difference as the $z$-coordinate to the points in the source map slice. The number of registration problems is the same as the consensus step in the proposed algorithm.

\subsubsection{ORB-TEASER++}

TEASER++ may incur a high memory and time cost, which was found to be the main bottleneck for the previous baseline. ORB-TEASER++ was tested as an attempt to reduce the number of registration steps to a minimum. In ORB-TEASER++, the feature matching step is performed between 2D keypoints across all of the slices in a map and the correspondences are used with TEASER++ for robust registration.

In essence, all 2D ORB features are treated as 3D features by assigning their relative height in the map's own frame as their $z$-coordinates. Then, the matching is performed against the 3D ORB features in the other map.

\subsubsection{FPFH-RANSAC}

ORB will be shown to not be an appropriate feature for 3D data in the tomographic manner, as it is prone to perceptual aliasing across different slices. For example, a single edge of a wall with no features would yield the same ORB signature for similar heights. Therefore, when the matching is performed with the points from the other map, many identical matches will exist. 

Instead, in FPFH-RANSAC the standard baseline of many point feature comparisons is used. FPFH \cite{Rusu2009FastPF}, features are computed for 3D Harris keypoints of the map point clouds, and a rigid transformation hypothesis is computed using RANSAC over point matches. 
An open-source implementation of this method \cite{horner2018automatic} is available online\footnote{\href{https://github.com/hrnr/map-merge}{\texttt{https://github.com/hrnr/map-merge}}}.

\subsubsection{FPFH-TEASER++}

As a modern alternative to RANSAC, which is known to be inaccurate under high outlier rates, this baseline uses TEASER++ for its registration task. The same features as FPFH-RANSAC baseline (FPFH descriptors of Harris 3D corners) are matched and registered using TEASER++ instead.

\subsubsection{DeepGlobalRegistration}

DeepGlobalRegistration \cite{choy2020deep} is the first learning-based baseline used. DeepGlobalRegistration uses FCGF \cite{Choy2019FullyCG}, assigns confidence for each correspondence, and solves a differentiable weighted Procrustes problem.

The complete pipeline is modified in the comparisons here to provide fair comparisons. Specifically, the ICP refinement and the safeguard registration implementations are deactivated to evaluate the network as-is, since these refinements are method-agnostic.

\subsubsection{GeoTransformer}

GeoTransformer \cite{qin2023geotransformer} adopts the transformer architecture \cite{Vaswani2017AttentionIA} to encode the geometric structure based on neighboring point geometries to obtain a pose-invariant superpoint representation. With high-inlier ratios of the GeoTransformer correspondences, accurate local-to-global pose estimation is made possible, even under low overlap scenarios.

\subsubsection{Rotation-Invariant Transformer (RoITr)}

RoITr \cite{yu2023rotation} proposes a cascade of attention modules that aims to eliminate any pose dependency on the features to isolate only the point geometry. In this manner, the architecture becomes inherently rotation-invariant.

RoITr is the only model among the others that do not provide model weights trained for KITTI dataset. For the real-data evaluation, RoITr is omitted.

\subsubsection{BUFFER}

BUFFER \cite{ao2023buffer} distills the point-based and patch-based feature extraction methods, improving point-wise orientation and saliency estimation and performs pose estimation via RANSAC over inlier set of correspondences as returned from the network.

\subsection{Comparison Metrics}

Performance of the map matching systems are evaluated based on four factors: a) translation error, b) rotation error, c) execution time, d) and memory usage.

\subsubsection{Translation Error}

Translation error is measured as the distance between the ground truth transformation estimation and the calculated estimation, and is reported in meters. Different applications require different voxel grid sizes. For example, an indoor robot with a small footprint will benefit from a higher resolution map (e.g. 0.05~m - 0.2~m grid size), whereas for an outdoor mission the same parameters will introduce large processing bottlenecks and yield a very large map, in terms of the occupied cells. To that end, the translation error threshold to determine if a map matching task is successful is assigned to be 5 times the grid size of the underlying map. This is in line with the literature for point cloud registration tasks for both indoor and outdoor scenarios.

\subsubsection{Rotation Error}

Rotation error is calculated from the difference in yaw angles between the ground truth transformation and the estimated transform, and is reported in radians. Unlike the translation error, the rotation error does not depend on the map resolution. The error threshold is set as 0.1745 radians ($\approx 10^\circ$), which is more conservative than the point cloud registration literature. The extents of the maps in map matching tasks are much larger in comparison and small variations can result in very large deviations.

\subsubsection{Execution Time}

Execution time is measured as the total time it takes the algorithm to compute the transformation parameters, reported in seconds. Any redundant pre-processing and post-processing steps are excluded from this measurement.

For the proposed method and baselines without a learning component, the computations were carried out on an Intel Phantom Canyon NUC11PHKi7C (Intel i7-1165G74 CPU, 64 GB RAM). Some deep learning-based pipelines required more memory than it is available on the aforementioned hardware, due to the scale of the problem. As such, the computations for learning-based baselines were carried out on a high-performance computer node with an NVIDIA A100 SXM4 80 GB card using 8 CPU cores and 530~GB RAM. While such hardware cannot be put on a mobile platform currently, this study takes into consideration the availability of better hardware in the future.

\subsubsection{Memory Usage}

Memory usage refers to the total amount of physical memory the algorithm utilizes to generate its result, in Bytes. Resident set size is selected as an estimate of the memory consumption for the proposed method and the first 4 baselines that do not use learning-based algorithms. For deep learning-based pipelines, the peak GPU memory usage is used as a measurement for the memory.

The following sections will introduce the simulation and real datasets and provide the results for the aforementioned baselines and the proposed method.

\section{Simulation Studies}\label{sec:simulation-data}

\begin{figure}[htbp]
    \begin{subfigure}{0.49\linewidth}
        \centering
        \includegraphics[keepaspectratio, width=\linewidth]{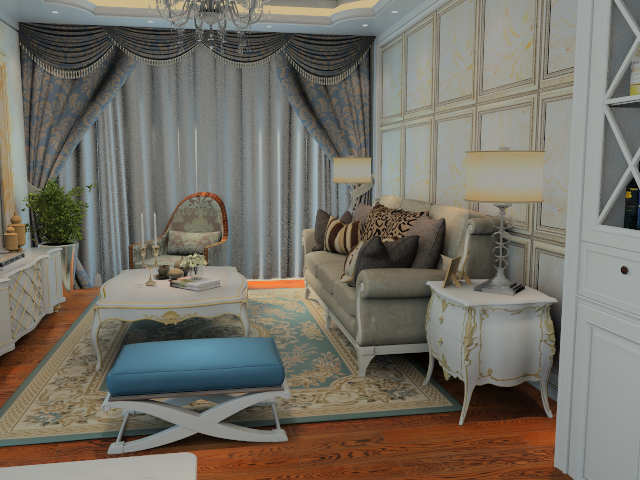}
        \caption{Color image}
    \end{subfigure}
    \hfill
    \begin{subfigure}{0.49\linewidth}
        \centering
        \includegraphics[keepaspectratio, width=\linewidth]{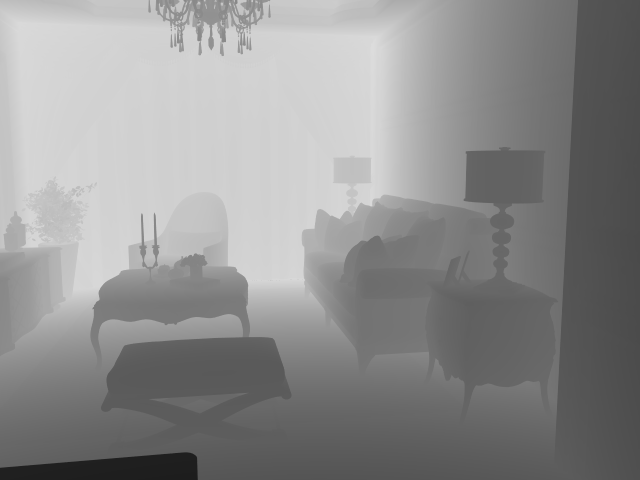}
        \caption{(Normalized) Depth image}
    \end{subfigure}
    \caption{A sample pair of color and depth images from the InteriorNet dataset.}
    \label{fig:interiornet-camera-depth}
\end{figure}

Simulation data is taken from individual trajectories of InteriorNet dataset \cite{Li2018InteriorNetMM}, in which a camera executes up to 3 different continuous 6 DoF motions in various furnished indoor environments. Color and depth streams, matched with the camera pose, are used to generate a 3D occupancy grid map to generate sample datasets for this study. A representative pair of color and depth images are provided in Figure~\ref{fig:interiornet-camera-depth}

The pose information for different trajectories in the same environment are provided with respect to a common, fixed frame of reference. In order to simulate a multi-agent mapping scenario, the initial pose of the trajectory is set as the origin of that trajectory, after removing the initial roll and pitch. In that way, each trajectory is emulating a different agent with its own local origin.

Five sequences from InteriorNet dataset were selected, each with three trajectories. In total, there are 30 different instances of map matching tasks (5 sequences $\times$ 3 trajectories $\times$ 2 directions). The average overlap between pairs of maps is calculated as 71.76\% (minimum 48.74\%, maximum 91.35\%).

The InteriorNet dataset provides noiseless measurements with perfect pose information. To test the robustness of the algorithms against common sources of noise, two different noise modalities were tested: pose estimation noise, and measurement noise.

\begin{figure}[htbp]
    \begin{subfigure}{0.3\linewidth}
        \centering
        \includegraphics[keepaspectratio, width=\linewidth]{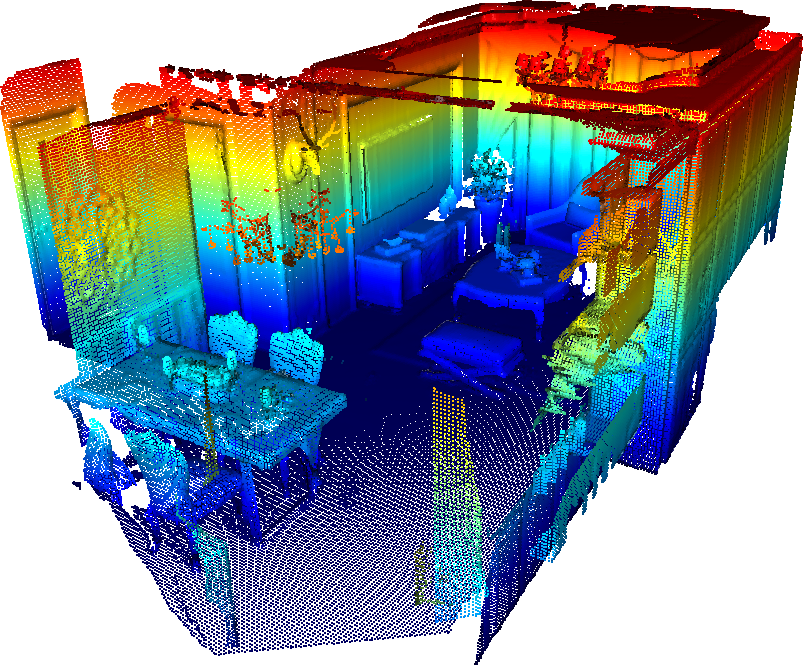}
        \caption{$\sigma = 0.00$}
    \end{subfigure}
    \hfill
    \begin{subfigure}{0.3\linewidth}
        \centering
        \includegraphics[keepaspectratio, width=\linewidth]{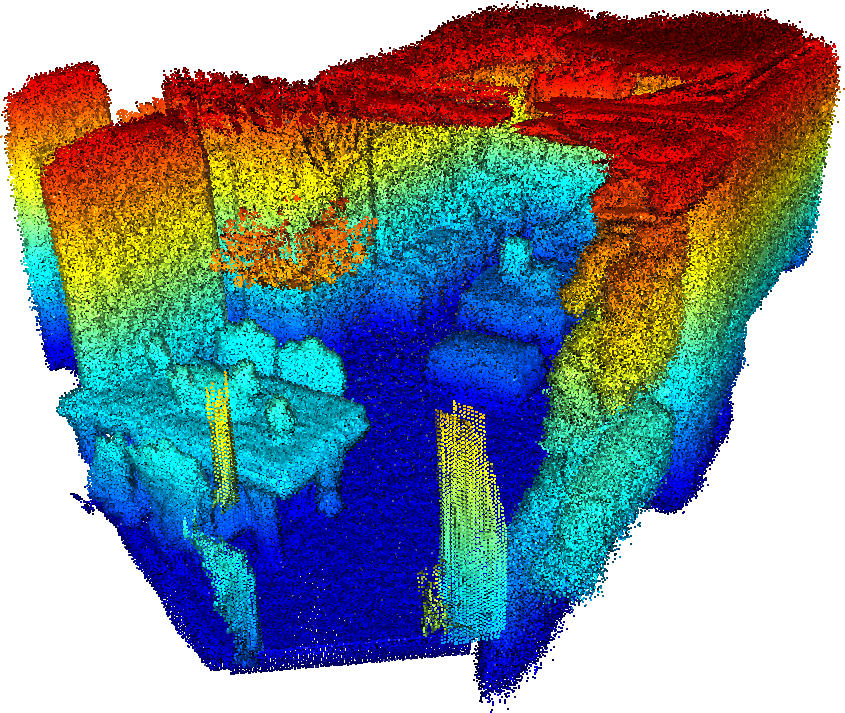}
        \caption{$\sigma = 0.02$}
    \end{subfigure}
    \hfill
    \begin{subfigure}{0.3\linewidth}
        \centering
        \includegraphics[keepaspectratio, width=\linewidth]{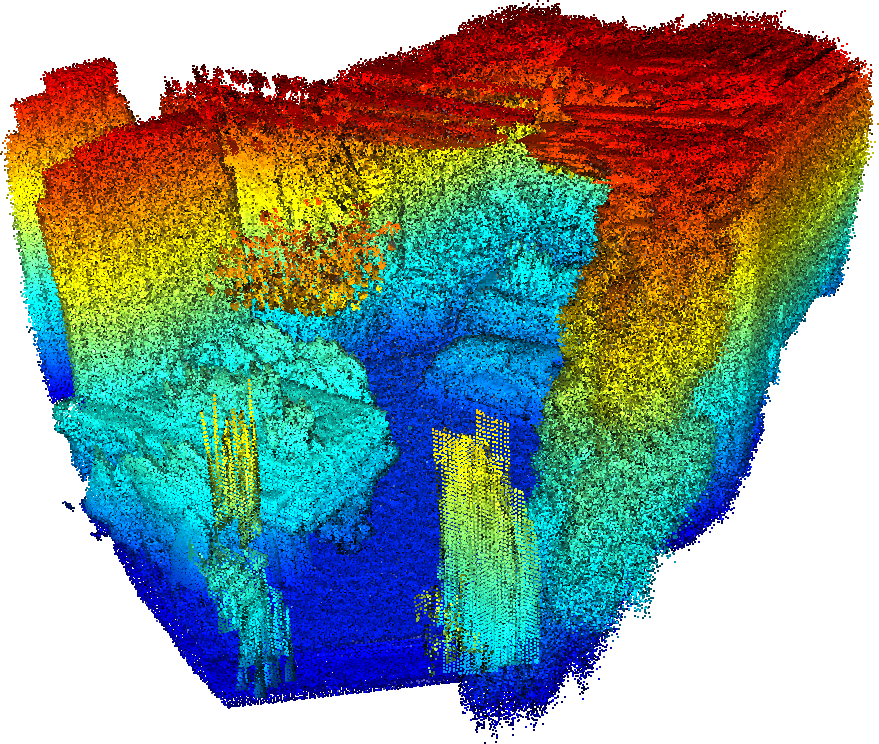}
        \caption{$\sigma = 0.05$}
    \end{subfigure}
    \caption{Sample point clouds from InteriorNet dataset under various noise conditions.}
    \label{fig:interiornet-sample-data}
\end{figure}

In order to simulate inaccuracies arising from noisy IMU measurements and imperfect mapping algorithm conditions, the ground truth pose for each camera reading is perturbed independently in each degree of freedom. Yaw, pitch, roll, and 3 translation perturbations are randomly sampled from independent, zero-mean Gaussian distributions with two different standard deviations. For $\sigma = 0.02$, the translation and rotation perturbations are sampled from distributions with standard deviations $0.02~$m and $0.01~$rad, respectively. For $\sigma = 0.05$, the standard deviations are $0.05~$m and $0.025~$rad. For the instances with added noise, the depth measurements from the camera were also corrupted with a noise profile mimicking the commercially available Intel RealSense D435i depth sensor. Some sample maps obtained under various noise levels are provided in Figure~\ref{fig:interiornet-sample-data} for a qualitative comparison.

\subsection{Map Matching Results}

\begin{figure}[ht]
    \centering
    \includegraphics[keepaspectratio, width=\linewidth, trim={0pt, 15pt, 0pt, 0pt}]{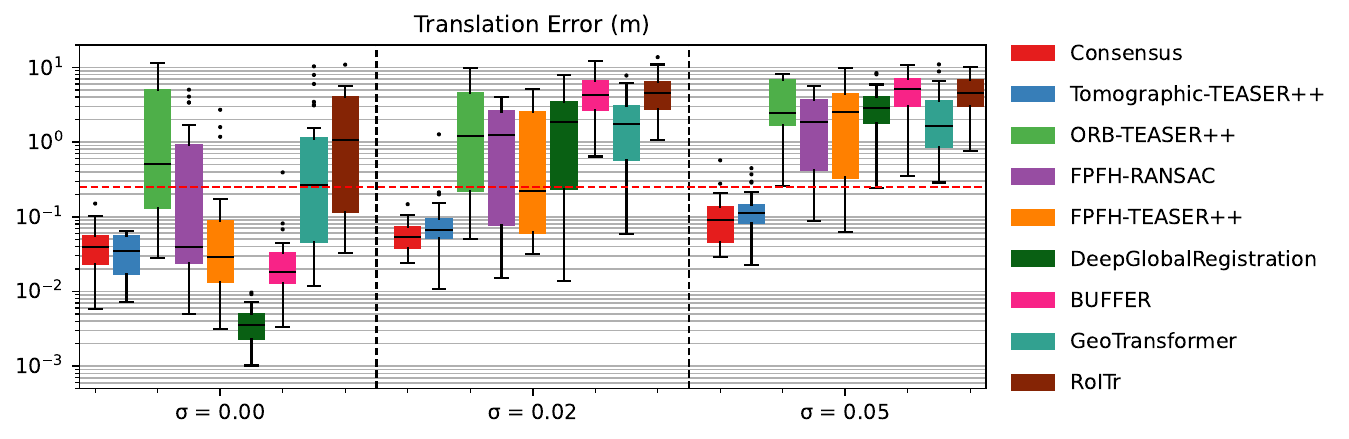}
    \caption{Translation errors of the tested algorithms for the InteriorNet data on various noise levels.}
    \label{fig:interiornet-dt}
\end{figure}

\begin{figure}[ht]
    \centering
    \includegraphics[keepaspectratio, width=\linewidth, trim={0pt, 15pt, 0pt, 0pt}]{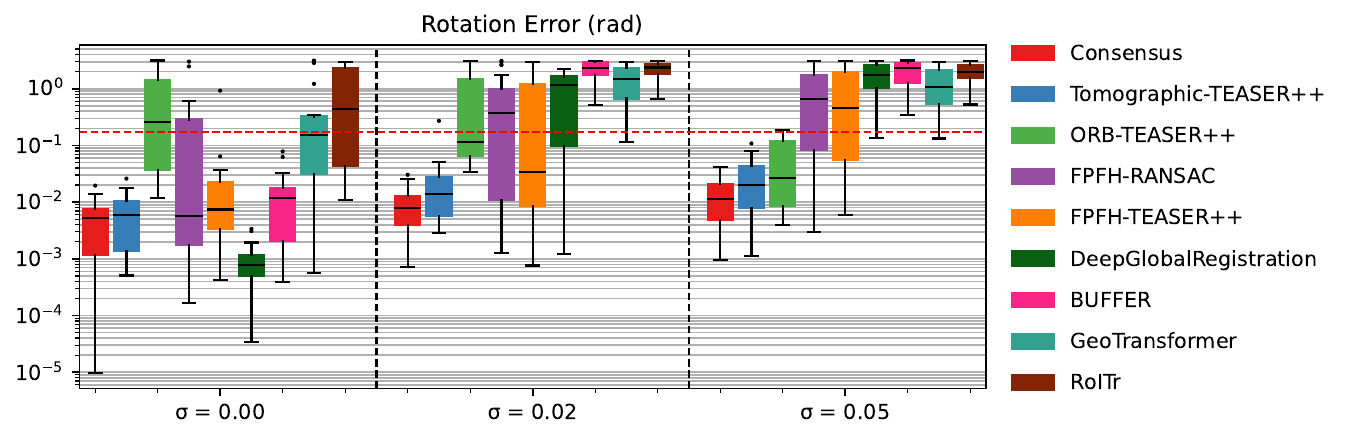}
    \caption{Rotation errors of the tested algorithms for the InteriorNet data on various noise levels.}
    \label{fig:interiornet-dr}
\end{figure}

\begin{figure}[ht]
    \centering
    \includegraphics[keepaspectratio, width=\linewidth, trim={0pt, 15pt, 0pt, 0pt}]{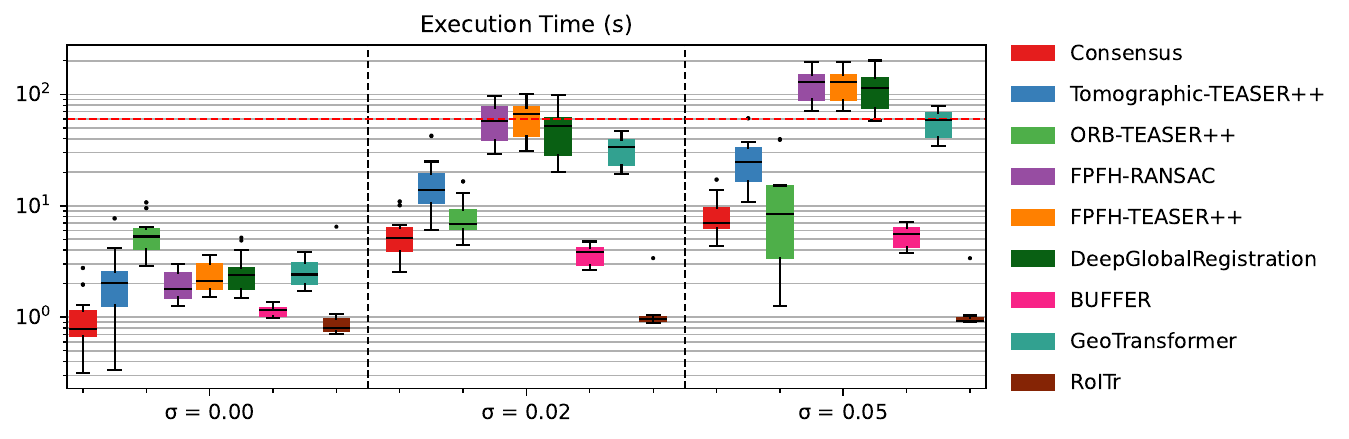}
    \caption{Execution time of the tested algorithms for the InteriorNet data on various noise levels.}
    \label{fig:interiornet-t}
\end{figure}

\begin{figure}[ht]
    \centering
    \includegraphics[keepaspectratio, width=\linewidth, trim={0pt, 15pt, 0pt, 0pt}]{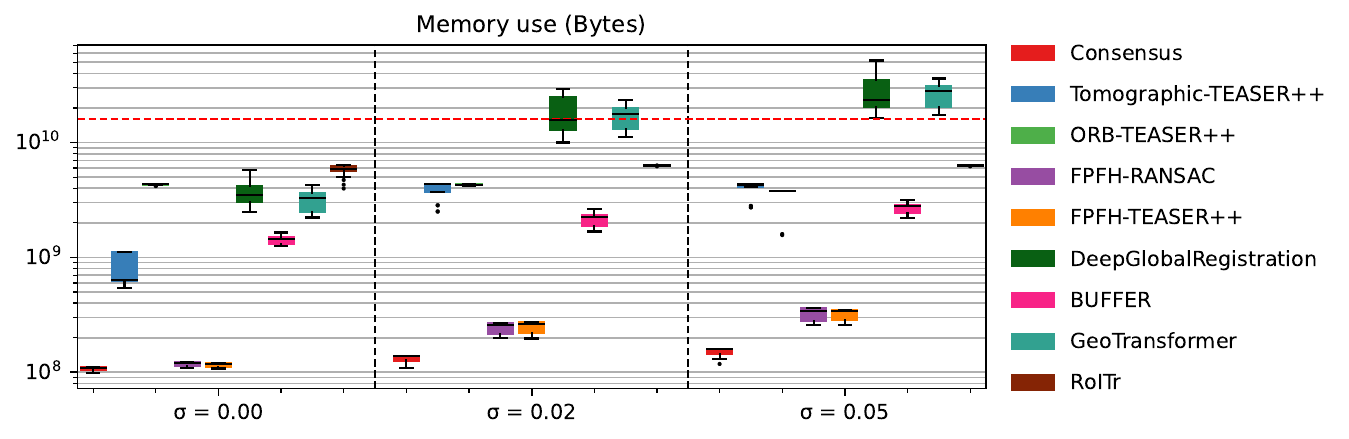}
    \caption{Memory usage of the tested algorithms for the InteriorNet data on various noise levels. For the first 5 algorithms, peak resident set size for CPU is reported. For the rest, peak GPU memory usage is reported.}
    \label{fig:interiornet-mem}
\end{figure}

Figures~\ref{fig:interiornet-dt}-\ref{fig:interiornet-mem} provide the translation and rotation errors, memory footprint, and execution times of each method on different levels of noise. The results are reported for maps that have been pre-filtered with a voxel grid filter of leaf size 0.05~m.

Each plot has a red dashed horizontal line, whose value represents a particular threshold. The selection is mostly subjective, success/failure classification is not performed  based on these thresholds.

Translation error threshold is set to 0.25~m, five times the grid size of the maps. Rotation error threshold is set to 0.1745~rad ($\approx 10^\circ$). Execution time threshold marks the 60~s line. Memory threshold is set to be 16~GB, which is the maximum capacity of the Jetson Orin NX computer, which can be attached to any mobile platform.

The noiseless case (leftmost panels in each figure) serves as an indicator that the baselines can actually handle the map matching task, especially for the learning-based methods. Consensus, Tomographic-TEASER++, FPFH-TEASER++, DeepGlobalRegistration (DGR), and BUFFER provide accurate results for the majority of the map pairs. Learning-based methods provide better results, with DGR being the most accurate. FPFH-TEASER++ shows a large variance with some failures. GeoTransformer and RoITr are still accurate for some pairs, but they fail to provide correct transform for the majority of the pairs. The failure of ORB-TEASER++ indicates that ORB features are not appropriate for use as a 3D feature descriptor, but their tomographic use (Tomographic-TEASER++) render them useful. 

All of the algorithms return with an estimate within 20 seconds. RoITr and Consensus methods take less than a second for the majority, with Consensus showing a larger variance and some outliers that reach up to 3 seconds.

For the memory use, however, the non-learning and non-TEASER++ methods shine. The memory use is limited to approximately 100~MB for Consensus and FPFH-based methods, at least an order of magnitude lower than any other method. The maximum clique search over the correspondence graph for TEASER++ has a large memory footprint for the ORB-TEASER++ implementation.

For the noisy data (middle and right panels in Figures~\ref{fig:interiornet-dt}-\ref{fig:interiornet-mem}), all of the learning-based methods fail. DGR provides some correct results, but on average is not reliable. FPFH features lose their discrimination capabilities under the noisy instances, as seen by the decreased accuracy between ideal and noisy instances. On the other hand, the tomographic methods demonstrate a more graceful accuracy degradation. The noise appears to be less impactful on the projected cross-sectional slices, when compared to other 3D descriptors.

The execution time increases with the noise in the input data, with the exception of RoITr. The execution time gap across methods is more pronounced at higher noise levels. Tomographic methods take longer to execute, as the noisy map means there is a larger number of slices across all maps. Consensus method remains approximately an order of magnitude faster in comparison to DGR and GeoTransformer. The other learning-based algorithms pass the 60 second mark. The bottleneck for FPFH-based methods becomes the computation of keypoints and the features and they cross the minute mark as well.

Finally, memory usage for all of the methods increase with the increased noise. Tomographic-TEASER++ reaches the correspondence cap utilizes the same amount of memory as the ORB-TEASER++. FPFH-based methods show limited increase, suggesting that the memory footprint is mostly dominated by the feature computation, rather than the robust pose estimation. The most memory-efficient method remains to be the Consensus, utilizing less than 200~MB on all instances.

Overall, tomographic methods appear to be the the better alternatives. In the ideal scenario, Tomographic-TEASER++ provides the smallest translation error. However, the Consensus method is more accurate in the noisy instances, and has the smallest execution time and memory footprint across all of the data types. The success of FPFH-TEASER++, DGR and BUFFER on noiseless data indicate that the algorithms can provide accurate results. However, the feature quality appears to degrade significantly under noisy data, resulting in quick accuracy degredation.

\section{Experimental Studies}
\label{sec:experimental_studies}

\subsection{KITTI Odometry Data}
\label{ssec:kitti-data}

\begin{figure}[ht]
    \begin{subfigure}{0.53\linewidth}
        \centering
        \includegraphics[keepaspectratio, width=\linewidth]{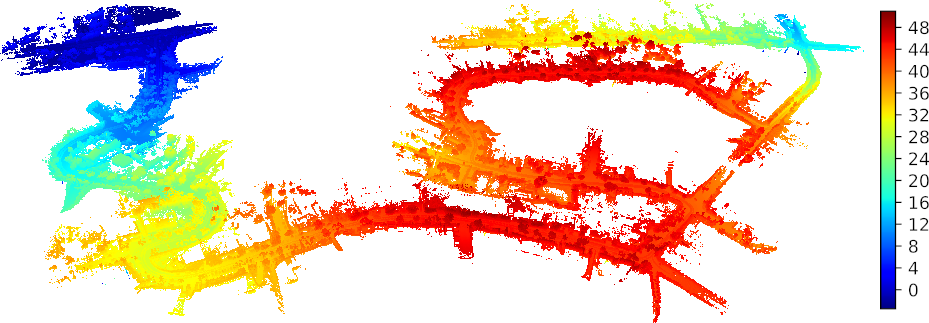}
        \caption{Map 1 - First half}
    \end{subfigure}
    \hfill
    \begin{subfigure}{0.45\linewidth}
        \centering
        \includegraphics[keepaspectratio, width=\linewidth]{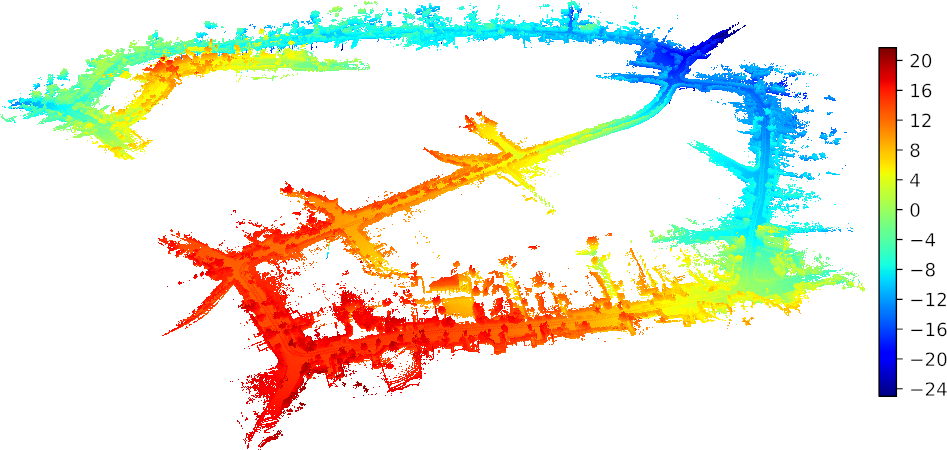}
        \caption{Map 2 - Second half}
    \end{subfigure}
    \caption{A sample map pair from KITTI dataset (sequence 02). The elevation is color-coded according to the scales on the right.}
    \label{fig:kitti-sample}
\end{figure}

To evaluate the performance on a real-life dataset, the LiDAR data from the sequences of KITTI odometry benchmark~\cite{Geiger2012CVPR} was used. In total, 11 sequences have GPS ground truth available. Out of the 11, only 5 sequences (00, 02, 05, 06, 07) revisit the previously explored locations and can be used for a map matching scenario. We divide the sequences in half to emulate two agents and perform matching for each agent, providing a total of 10 different instances of map merging. Average overlap was estimated at 43.90\% (minimum 11.17\%, maximum 83.53\%). Figure~\ref{fig:kitti-sample} provides an example pair for the map matching task.

For this study, two of the learning-based baselines could not be used. RoITr model does not provide any weights for KITTI or hyperparameter set that can be used to train the architecture on KITTI sequences. For GeoTransformer, the data preparation step filled up the RAM of the high-performance computing node (approx. 530~GB) and could not be run.

\begin{figure}[htb]
    \centering
    \includegraphics[keepaspectratio, width=\linewidth, trim={0pt, 15pt, 0pt, 0pt}]{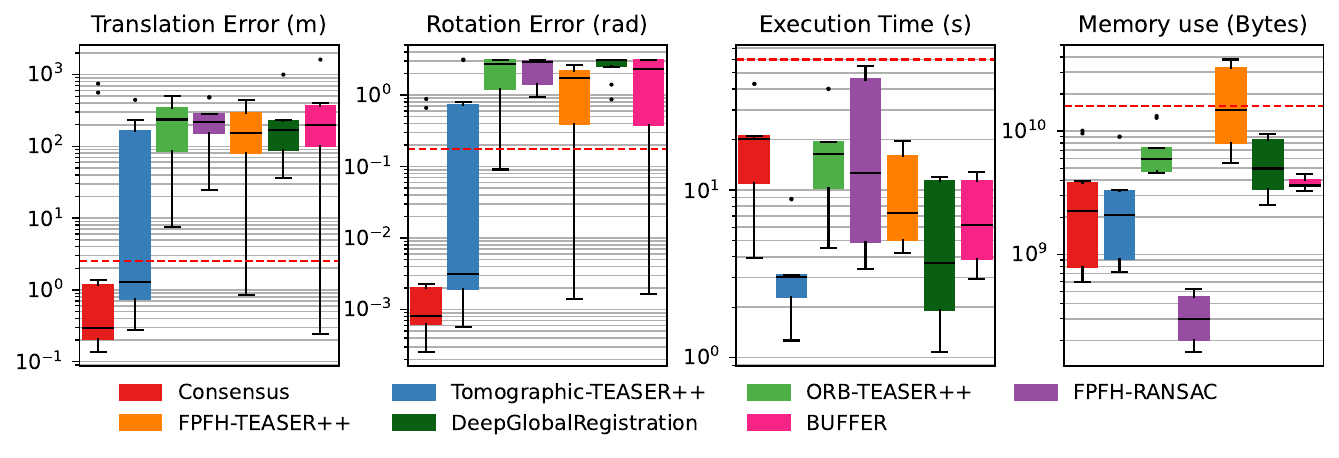}
    \caption{Aggregated errors, execution time, and the memory usage of the tested algorithms on KITTI odometry dataset, pre-filtered to a grid size of 0.50~m.}
    \label{fig:kitti-g0.50}
\end{figure}

\begin{figure}[htb]
    \centering
    \includegraphics[keepaspectratio, width=\linewidth, trim={0pt, 15pt, 0pt, 0pt}]{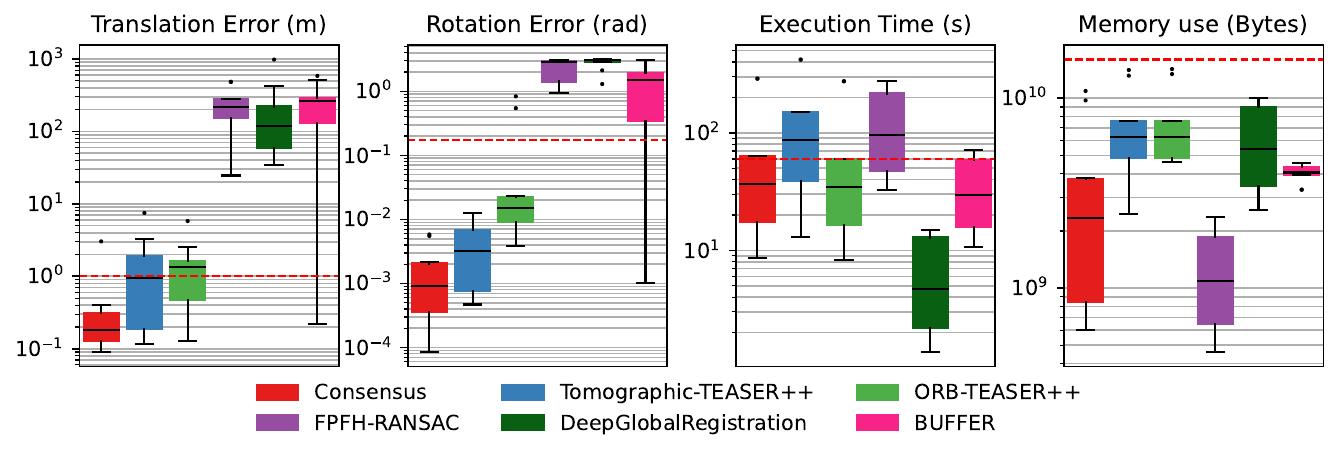}
    \caption{Aggregated errors, execution time, and the memory usage of the tested algorithms on KITTI odometry dataset, pre-filtered to a grid size of 0.20~m.}
    \label{fig:kitti-g0.20}
\end{figure}

Two grid sizes were used for the remaining baselines. Figure~\ref{fig:kitti-g0.50} reports the results for the data where the maps have been pre-filtered with a voxel grid filter of size 0.50~m. Results provided in Figure~\ref{fig:kitti-g0.20} use data that was pre-filtered to 0.20~m. Horizontal red lines mark the same thresholds as the InteriorNet case: five times the grid size for translation error, 0.1745~rad for rotation error, 60~s for the execution time, and 16~GB for the memory usage.

Due to the large number of correspondences, FPFH-TEASER++ ran out of memory in the mobile computer (64~GB) at higher resolution. As such, the method is omitted for the 0.20~m grid size (Figure~\ref{fig:kitti-g0.20}).

The only algorithm that is successful for the majority of the pairs is the tomographic methods. Consensus provides high accuracy for both rotation and translation. Tomographic-TEASER++ fails to estimate the translation accurately, although the rotation estimation is mostly succesful. Even though DGR was accurate in the simulated data, it fails to perform in the KITTI study. BUFFER is still able to provide the correct transformation for some of the instances, but overall is inaccurate. FPFH features appear to be not descriptive on the KITTI dataset, which could be due to improper parameter selection. 

The execution time is significantly higher for Consensus in comparison to the indoor data. For the higher resolution map, the execution time reaches up to a minute. The primary reason for the losses in efficiency is due to the much larger number of slices across both maps. Specifically for the sequence 02, which is shown in Figure~\ref{fig:kitti-sample}, the maps span approximately 50 meters. Slice correlation step is much more time consuming, and it hampers the time efficiency.

In a similar fashion, the memory footprint of the Consensus algorithm is significantly higher, reaching up to 10~GB for some instances of high-resolution map. The footprint remains lower than any other algorithm that performs accurate estimations. 

Overall, even though the Consensus algorithm efficiency and memory usage has increased significantly, it is still the best performing option out of the baseline methods.

Note that across the studies, we are only changing the grid size of the maps for the tomographic algorithms. Despite the minimal parameterization, the algorithm performance is not impacted significantly. The appropriate grid size selection is still a factor to be taken into consideration, but the choices here (0.2~m and 0.5~m) is common for LiDAR-based outdoor mapping tasks.

\subsection{Large-scale Indoor Environment Data}

\begin{figure}[ht]
    \begin{subfigure}{0.59\linewidth}
        \centering
        \includegraphics[keepaspectratio, width=\linewidth]{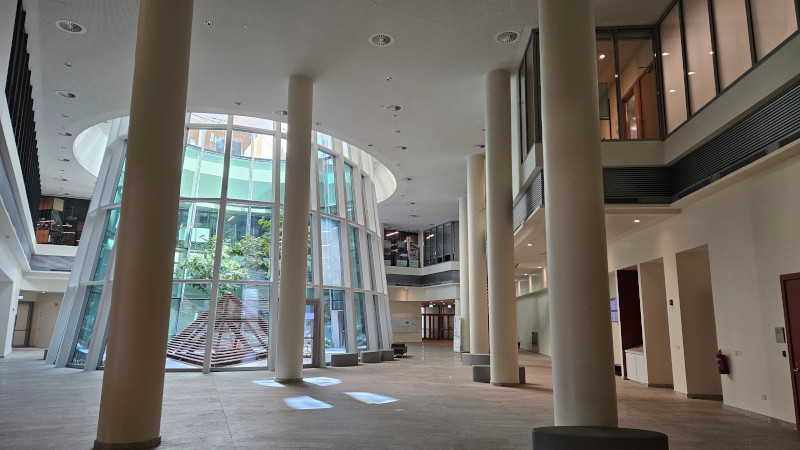}
        \caption{Large indoor environment for the experimental studies}
        \label{fig:a1-photo}
    \end{subfigure}
    \hfill
    \begin{subfigure}{0.4\linewidth}
        \centering
        \includegraphics[keepaspectratio, width=\linewidth]{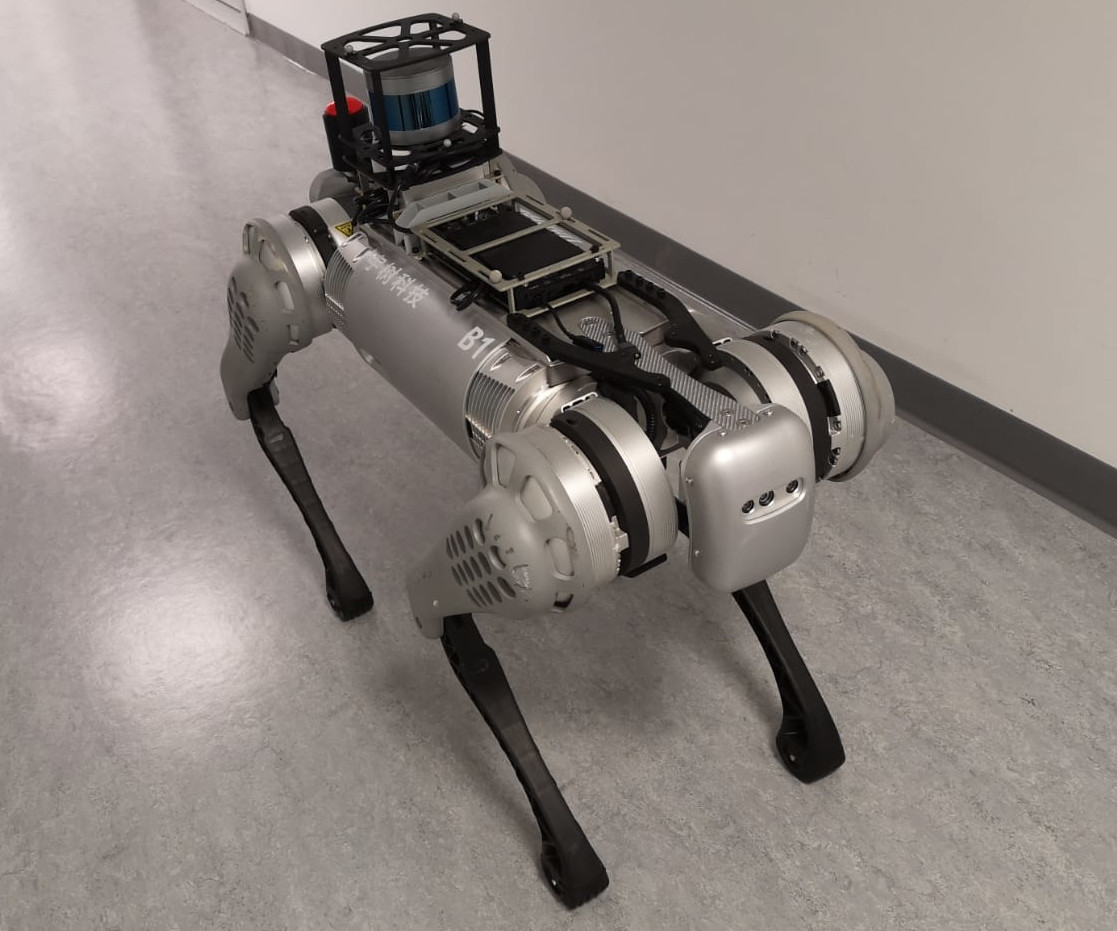}
        \caption{Unitree B1}
        \label{fig:robot}
    \end{subfigure}
    \caption{Photos of the environment (left) and the robotic platform (right)}
    \label{fig:experimental-photos}
\end{figure}

\begin{figure}[htbp]
    \centering
    \includegraphics[keepaspectratio, width=0.95\linewidth]{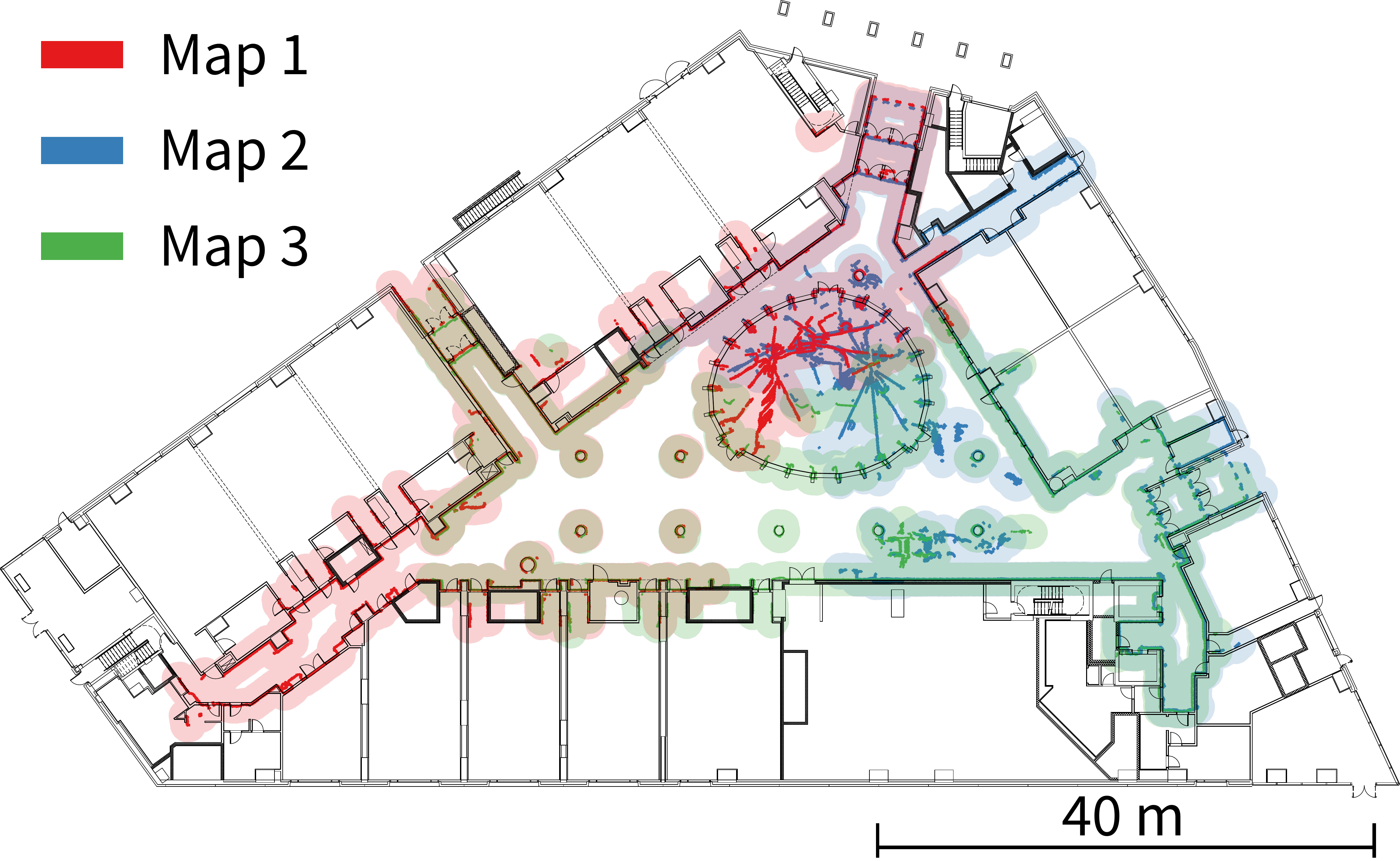}
    \caption{Individual maps overlaid on the architectural plan of the environment used in the experimental study. Each color indicates the area covered by an agent.}
    \label{fig:a1-layout}
\end{figure}

Finally, the algorithms' performance is evaluated using a generated dataset, mapping a large-scale indoor environment. The interior space is an atrium area connecting office spaces, classrooms, and laboratories across two floors. The environment is especially challenging for LiDAR-based distance sensors, as there is a glass construction at the center with opening to the outside. A photograph of the environment is provided in Figure~\ref{fig:a1-photo}. A Unitree quadruped robot equipped with RoboSense 3D LiDAR sensor (Figure~\ref{fig:robot}), was used to map the environment in 3 separate trajectories. The horizontal cross-sections of the individual maps overlaid on the architectural plan of the structure is provided in Figure~\ref{fig:a1-layout}. Each trajectory is mapping along two corners of the triangle.

Horizontal red lines mark the same thresholds as the InteriorNet case: five times the grid size for translation error, 0.1745~rad for rotation error, 60~s for the execution time, and 16~GB for the memory usage.

To demonstrate the applicability to different forms of mapping tools, the robot estimates its odometry using KISS-ICP~\cite{vizzo2023kiss} and generates the map using VDBFusion~\cite{vizzo2022vdbfusion} framework. KISS-ICP estimates relative transformations between two key frames using point-to-point ICP with simple motion compensation, an adaptive correspondence threshold, and a robust kernel to provide a minimally-parameterized solution. VDBFusion represents the environment as a truncated signed distance function (TSDF) to obtain more accurate dense reconstructions, powered by OpenVDB\footnote{\href{https://www.openvdb.org/}{\texttt{https://www.openvdb.org/}}}, an open-source, production-level hierarchical data structure library. In this way, we demonstrate backend-agnostic capabilities of our method in the map matching task.

\begin{figure}[htbp]
    \centering
    \begin{subfigure}{0.5\linewidth}
        \centering
        \includegraphics[keepaspectratio, width=\linewidth]{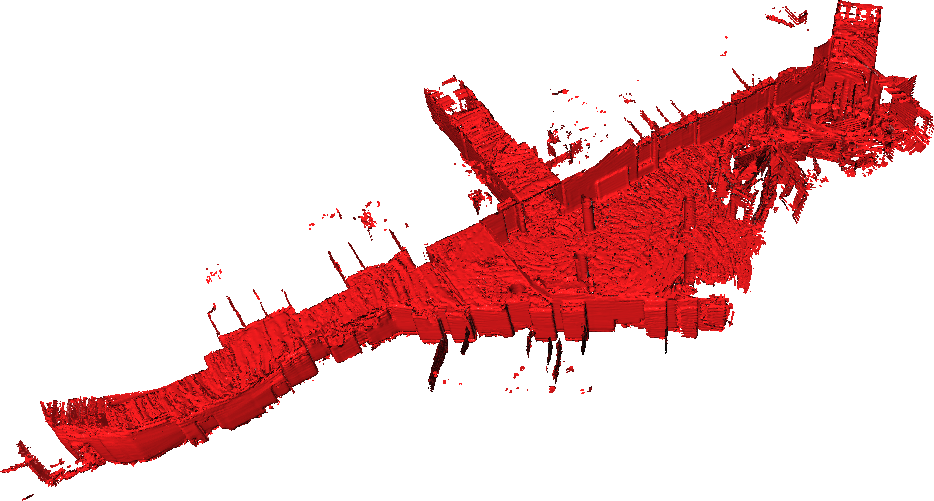}
        \caption{Map 1: 2,290,006 points}
    \end{subfigure}
    \hfill
    \begin{subfigure}{0.48\linewidth}
        \centering
        \includegraphics[keepaspectratio, width=\linewidth]{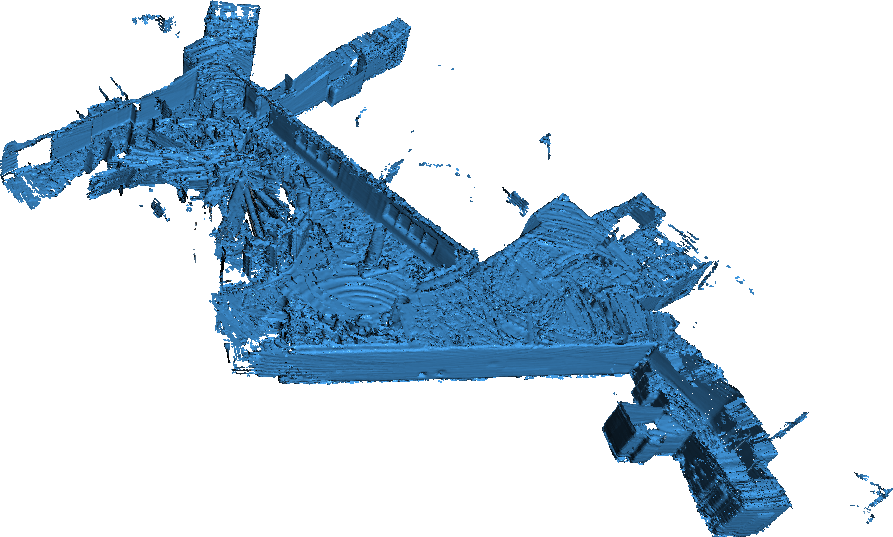}
        \caption{Map 2: 2,205,667 points}
    \end{subfigure}
    \begin{subfigure}{0.7\linewidth}
        \vspace{8pt}
        \centering
        \includegraphics[keepaspectratio, width=\linewidth]{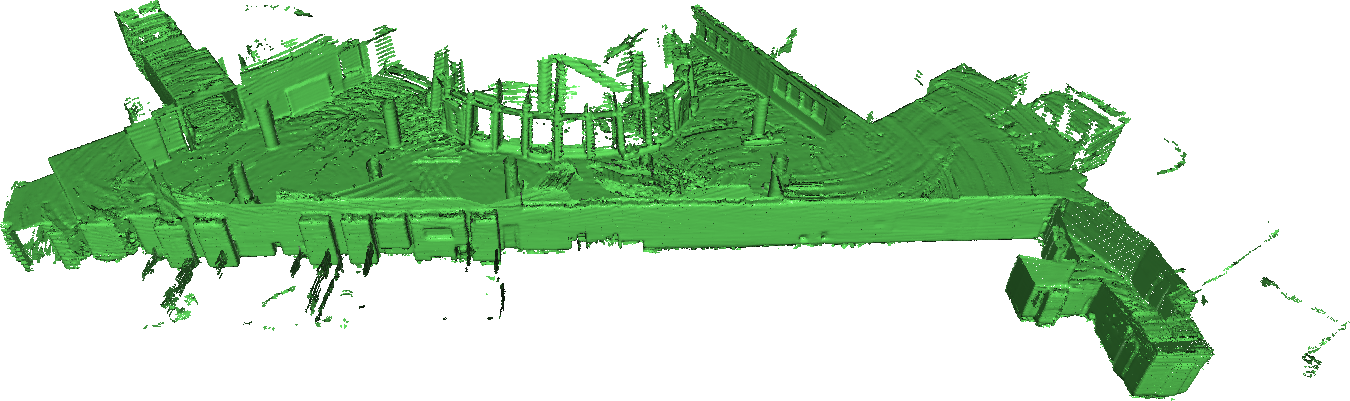}
        \caption{Map 3: 1,651,985 points}
    \end{subfigure}
    \caption{Pose-accurate renderings and the number of points for each map used in the experimental study. The colors match the overlays from Figure~\ref{fig:a1-layout}.
    }
    \label{fig:a1-maps}
\end{figure}

To generate the map, deskewed LiDAR point cloud data is integrated into the VDBFusion map, and the 3D triangle mesh representation of the resultant map was extracted using marching cubes algorithm at a grid size of 0.05~m. Vertices of the mesh are provided as input to all of the maps. A rendering of the the three maps generated as described above are provided in Figure~\ref{fig:a1-maps}.

\begin{figure}[htbp]
    \centering
    \includegraphics[keepaspectratio, width=\linewidth, trim={0pt, 15pt, 0pt, 0pt}]{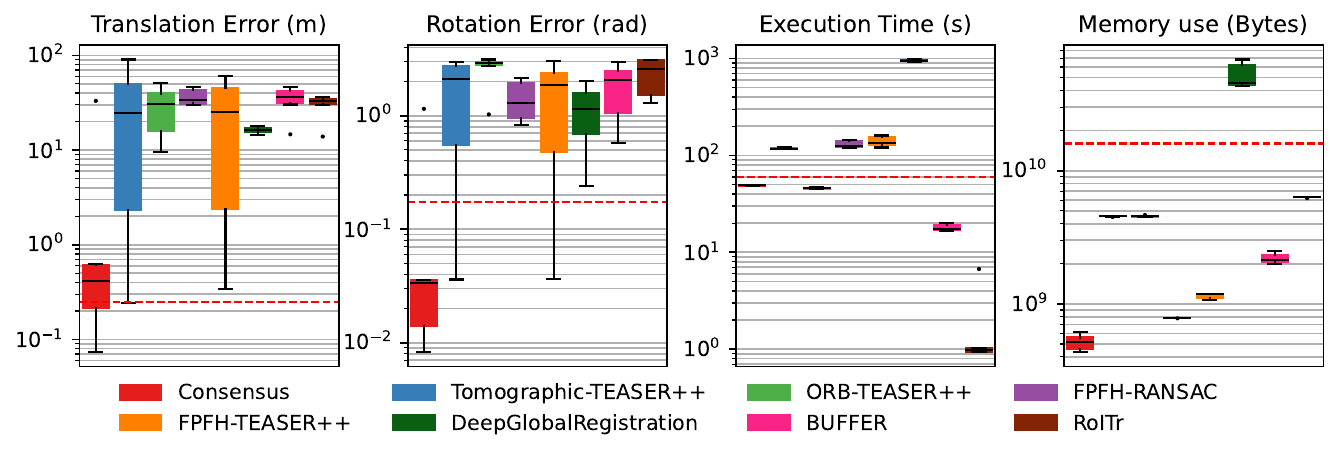}
    \caption{Aggregated errors, execution time, and the memory usage of the tested algorithms for the experimental study. Horizontal line for the translation error is at 0.25~m.}
    \label{fig:a1-results}
\end{figure}

Overall, three maps and two directions provide a total of 6 map matching tasks. Similar to the KITTI study, Geo-Transformer baseline could not be run due to memory consumption. The results for all of the other methods are provided in Figure~\ref{fig:a1-results}.

Only the estimates provided by the proposed Consensus method are close to the true transformations. Tomographic TEASER++ and FPFH-TEASER++ provide accurate orientation for some of the pairings, but overall no other algorithm is able to provide accurate results.

Even for the Consensus algorithm, the translation error is much larger. The algorithm is able to estimate the correct rotation for all but one pairing. A parameter tuning may improve the translation error as well, but the finer registration with ICP is always possible. The algorithm is providing reasonable global transformation estimates.

Execution time is reaching to the minute mark for the proposed method in this framework. The grid size is small enough to generate both large-size slices for feature extraction and a large number of slices to be cross-correlated. However, the memory footprint is not affected, utilizing approximately 500~MB in RAM to carry out the computations.

\section{Limitations}\label{sec:limitations}

Our proposed algorithm has inherent limitations that render it unusable for generic point cloud registration tasks. The three most important ones are the missing degrees of freedom, the inability to deal with non-rigid estimations, and the dependence on the map heights.

Our assumption is that the agents have accurate gravity estimates that they can use to align their maps. IMU sensors are common, but are not entirely accurate. Large accelerations can affect the orientation quality and impact the overall map alignment. Even though the Consensus method was demonstrated to be  resilient to noise, there is no way to recover the roll and pitch if the origin (the initial pose) of the agent is misaligned. 

Another limitation is that the output is a single rigid transformation. In practice, large-scale maps tend to drift even if robust mapping methods are used. As such, a single rigid transformation is generally not the most accurate answer. A deformable hypothesis operating on the pose graph \cite{bonanni20173} would me more appropriate, and there is no way to adapt this method directly.

Finally, the complexity of the algorithm depends mostly on the number of slices for each map. Per-slice transform estimations are fast, but the correlation step requires many hypotheses to be tested at once. The higher the number of slices, the heavier the computations. Hardware-accelerated parallelization could sidestep the computational bottlenecks, but the dependence on the map heights will not change.

Furthermore, the comparative studies, performed against the baselines has certain limits. These limitations can be grouped into 3 categories: parameter tuning for the tasks, disparity in the execution paradigms, and potentially suboptimal or inaccurate adaptations.

The weights used for learning-based models are taken directly from the best parameters as provided by the original authors. For the simulated study, this implies training on 3DMatch \cite{Zeng20163DMatchLL} dataset, which contains 3D scans of small-scale scenes. Even though the grid size parameter is appropriate, we do not account for the simulation-to-real domain gap, and the scale of the problem is not the same. The results on the ideal data (Figures~\ref{fig:interiornet-dt}-\ref{fig:interiornet-mem}, left-most panels) indicates that some of the algorithms can still perform, but a much thorough comparison would require to train the models on data that is more representative for the type and the scale of the map matching task.

For the FPFH-based methods, an extensive search on the key parameters, mainly the radii to estimate normals, keypoints, and features, was not performed. In a similar fashion to the tomographic methods, the radii was scaled based on the grid size, but a parameter study could yield a more optimal answer. The only parameter that was changed across different datasets is the grid size of the maps, which is known ahead of the time. The proposed algorithm is able to perform accurately with minimal parameterization. As such, the advantage of tomographic methods remain.

The efficiency is compared using the execution time and memory footprints. An ideal scenario would require the programming paradigms and the hardware platforms to be the same. However, learning-based methods are implemented primarily on Python, whereas we implemented our method and the remaining baselines (Consensus, ORB-family and FPFH-family of algorithms) in C++. Even though the Python bindings for the learning-based algorithm backends are compiled as well, the comparison is not entirely applicable. Furthermore, Python memory management is different than C++, introducing additional sources of issue.

On the other hand, the metrics were calculated for learning-based algorithms on a high-performance computer node that is equipped with a powerful graphics card. While we attempted to provide the best possible comparison, this limitation should be explicitly stated.

Finally, not all of the learning-based baselines had an existing implementation to perform registration on arbitrary datasets. The implementations from BUFFER and RoITr were mimicked from the source code to obtain the transformations.

\section{Conclusions}\label{sec:conclusion}

Tomographic features were proposed to address large-scale map matching task in gravity-aligned 3D maps. The proposed minimally-parameterized approach was both accurate and efficient in memory and computation power. Furthermore, the tomographic approach was found to be more resilient to measurement noise in sensors in comparison to other methods utilizing 3D feature descriptors, learning-based or otherwise. The findings are corroborated on real datasets that map volumes of different scales, underscoring the algorithmic efficiency and accuracy.

The proposed algorithm cannot provide a non-rigid estimate to deal with mapping inaccuracies, and is not able to estimate the roll and pitch. Nevertheless, for large and dense maps with bounded extents, such as the case in many graph-based SLAM algorithms, the tomographic approach can provide fast and robust transformations and requires investigations.

\bibliographystyle{elsarticle-num-names}
\bibliography{references}

\section*{Author Contributions}

Co-author contributions follow the CRediT taxonomy:
\textbf{Halil Utku Unlu:} Conceptualization, Methodology, Software, Validation, Formal analysis, Investigation, Data curation, Writing - original draft, Writing - review and editing, Visualization.
\textbf{Anthony Tzes:} Conceptualization, Methodology, Resources, Writing - review and editing, Supervision.
\textbf{Prashanth Krishnamurthy:} Conceptualization, Methodology, Writing - review and editing.
\textbf{Farshad Khorrami:} Conceptualization, Methodology, Writing - review and editing, Supervision.

\section*{Funding}

This work was partially supported by the NYUAD Center for Artificial Intelligence and Robotics (CAIR), funded by Tamkeen under the NYUAD Research Institute Award CG010. The funding sources were not involved in study design, in the collection, analysis and interpretation of data, in the writing of the report, and in the decision to submit the article for publication.

\end{document}